\batchmode
\makeatletter
\def\input@path{{\string"D:/MP/RECHERCHE/article-2012-Human-Computer Studies/\string"/}}
\makeatother
\documentclass[twoside,english]{elsarticle}
\usepackage[latin9]{inputenc}
\usepackage{color}
\usepackage{amsthm}
\usepackage{amstext}
\usepackage{graphicx}

\makeatletter

\newcommand{\lyxmathsym}[1]{\ifmmode\begingroup\def\b@ld{bold}
  \text{\ifx\math@version\b@ld\bfseries\fi#1}\endgroup\else#1\fi}

\providecommand{\tabularnewline}{\\}

\theoremstyle{plain}
\newtheorem{thm}{\protect\theoremname}
\theoremstyle{definition}
\newtheorem{defn}[thm]{\protect\definitionname}

\journal{International Journal of Human-Computer Studies}

\makeatother

\usepackage{babel}
\providecommand{\definitionname}{Definition}
\providecommand{\theoremname}{Theorem}

\begin{document}

\title{Visualizing and Interacting with Concept Hierarchies}

\author{Michel~Crampes\corref{cor1} }

\ead{michel.crampes@mines-ales.fr}

\ead[url]{http://www.lgi2p.ema.fr}

\author{Michel~Plantié\corref{cor2}}

\ead{michel.plantie@mines-ales.fr}

\cortext[cor1]{Principal Corresponding author}

\cortext[cor2]{Corresponding author}

\address{Laboratoire LGI2P, Ecole des Mines d'Ales, \linebreak{}
site EERIE, Parc George Besse, 30000 N\^{i}mes, France}
\begin{abstract}
Concept Hierarchies and Formal Concept Analysis are theoretically
well grounded and largely experimented methods. They rely on line
diagrams called Galois lattices for visualizing and analysing object-attribute
sets. Galois lattices are visually seducing and conceptually rich
for experts. However they present important drawbacks due to their
concept oriented overall structure: analysing what they show is difficult
for non experts, navigation is cumbersome, interaction is poor, and
scalability is a deep bottleneck for visual interpretation even for
experts. 

In this paper we introduce semantic probes as a means to overcome
many of these problems and extend usability and application possibilities
of traditional FCA visualization methods. Semantic probes are visual
user centred objects which extract and organize reduced Galois sub-hierarchies.
They are simpler, clearer, and they provide a better navigation support
through a rich set of interaction possibilities. Since probe driven
sub-hierarchies are limited to users\textquoteright{} focus, scalability
is under control and interpretation is facilitated. After some successful
experiments, several applications are being developed with the remaining
problem of finding a compromise between simplicity and conceptual
expressivity.\end{abstract}
\maketitle

\section{Introduction}

Visualization and interaction are two major supports for searching
and analysing object sets. A lot of methods and tools have been proposed
to organize, represent and display objects for providing users with
immediate access to subsets of objects according to users\textquoteright{}
intention or some internal logic. But when the size of sets is important,
most of visual displays become difficult to interpret and interaction
turns into complex manipulation. Scalability is a serious bottleneck.
This is a paradox because visualization loses efficiency with complex
sets where it is expected to provide solutions for managing complexity
(\citet{Chen2005}). As a result most popular solutions for searching
objects of data collections present query results as lists of items,
such as Google, or grids of objects, such as Flickr or Facebook with
photos. It seems that sophisticated visualization solutions are for
experts and straightforward visualisation is for non-expert audiences.
This paper tackles the difficult problem of turning an expert visual
display to an interesting simple application for novices. 

Concept Hierarchies (CH) and Formal Concept Analysis (FCA) (\citet{GanterB.1999})
are examples of such methods which are particularly well grounded
on a theoretical point of view, largely experimented in numerous lab
applications, and, thanks to Galois lattices and line diagrams called
Hasse diagrams, particularly adapted to searching and analysing sets
of objects endowed with attributes. However Galois lattices still
fall short of managing visual complexity even for medium CHs (\citet{GanterB.1999,springerlink:10.1007/11671404_1,roth2006towards}):
\textquotedblleft{}Representing concept lattices constructed from
large contexts often results in heavy, complex diagrams that can be
impractical to handle and, eventually, to make sense of\textquotedblright{}
(\citet{Kuznetsov2007}). 

In this paper we address scalability and expressivity of concept hierarchies
for non-expert audiences. A new visualization and interaction paradigm
is presented with its key concept: user centred Semantic Probes. Visual
results are compared to a traditional Galois lattice and to the proposed
Galois Lattice reduction methods. It has been tested in controlled
experiments on a bechmark of 127 objects tagged with 245 attributes
and on real data, photo albums extracted from Facebook. It has raised
the interest of several industrials for which different applications
are being developed.

\begin{figure}[h]
\centering{}\includegraphics[width=7cm]{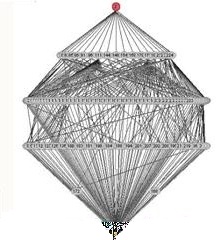}\caption{\label{figure1}A semantic probe driven Galois sub-hierarchy lattice}
\end{figure}

\section{RELATED WORKS AND MOTIVATION\label{section2}}

Visualizing sets of entities and their properties or relations, such
as biological data, multimedia objects or social activity, is an increasingly
important issue. The goal is to visually elicit known or hidden organization
that mere lists cannot reveal with the intention of extracting specific
knowledge or particular objects. A myriad of solutions have been proposed
which depend upon the designers\textquoteright{} intention and the
type of entities to visualize such as object-object relations (graphs)
(\citet{BattistaG.1999,Herman2000}) or multivariate data (\citet{Buja01visualizationmethodology,Kohonen:2001:SM:558021,Platt2005}).
In these fields, scalability is often a difficult problem. For example,
in graph visualization it is necessary to visualize hundreds and even
thousands of usually entangled links between objects. Different strategies
have been proposed such as clustering (\citet{Fortunato2009,Noack2008a,Holten2006,Noack2005}),
interaction techniques (panning, zooming, focus+context, filtering,
animation \citet{Herman2000,Shneiderman1996,Dwyer2008,Lamping:1995:FTB:223904.223956}),
dynamic local views centred on user\textquoteright{}s interest (\citet{Alani2003,VanHam2009}),
or multi-views for linking different complementary views (\citet{Streit2009}). 

In this paper we consider a different kind of entities, object-attribute
databases, which can be found in many areas where objects of some
type are tagged with attributes of a different type (e.g. photos with
tags, genes linked to their properties, etc.). The problem of displaying
and exploring their structure shares with graphs the same difficulty
of scalability. But it is even more challenging because relations
between objects are linked to attribute ownership which should consequently
be visually revealed. 

Formally object-attribute sets are equivalent to bipartite graphs.
They are graphs whose node set can be partitioned into two disjointed
sub-sets, and edges only link nodes from a sub-set to the other. In
our case, attribute set and object set are the two sub-sets and links
between two nodes represent attribute ownership. Many techniques for
visualising bipartite graphs have been proposed mostly focusing upon
avoiding as much as possible edge crossings for better interpretation.
The most notorious is the two-layer layout with its barycentre method
for minimizing edge crossings (\citet{BattistaG.1999}). However for
big bipartite graphs the visual result is still intricate. Recent
solutions make use of interaction techniques such as focus+context,
Fish Eye and information hiding to handle big data sets (\citet{schulz:hal-00656214}).
The authors argue that the resulting display is usable for experts,
but it is far from being simple and straightforward for novices. 

Other methods try to catch object-attribute data through node clustering
(\citet{Fluit2005}) and hypergraphs. In the last case nodes of one
of the two sets become hyper-edges containing the corresponding nodes
of the other set. With this respect matrices in (\citet{Riche2010})
or Euler diagram boxes in (\citet{Riche2010a}) are used to build
hyper-edges. Node duplication is analyzed in both papers to represent
hyper-edge intersections. However object duplication does not prove
to be visually the most appropriate solution for users in both papers.
Testers favour what the authors call Compact Rectangular Euler Diagrams
(\citet{Riche2010a}) where objects have unique visible identities.
In this respect underground maps are an interesting hypergraph metaphor.
Lines represent hyper-edges and stations stand for nodes which may
belong to several lines (\citet{Brandes2010}). This original technique
still needs experimentation with users to prove its interest. 

But in all these above visualization strategies, examples are based
on very small data sets and even under that limitation, interpretation
is still difficult. Whatever the method, drawing hypergraphs and Euler
diagrams is particularly cumbersome on limited data sets and even
more on real applications which require the display of large databases. 

The most common and formally fruitful approach for object-attribute
data visualization is based on Galois lattices (\citet{GanterB.1999,eklund:inria-00503254})
which are visualized through layered graphs called Hasse diagrams;
an example is given in Figure \ref{figure1}. Each node is identified
by a subset of objects and a subset of attributes; edges link nodes
according to a partial order relation. The detailed state of the art
in this domain will be presented after introducing FCA basics in the
next section. 

The four visualization methods (two-layer layouts, matrices, Euler
diagrams and Hasse diagrams) have been deeply studied by researchers
with many variations. But usability is still questionable because
we see few everyday applications of these methods. Conversely when
searching object-attribute databases most common applications display
query results as lists (i.e. Google) or grids (i.e. Flickr or Facebook)
of objects. Objects may be ordered according to their proximity with
the query, but little information is given about the semantics, the
ordering, or the structure of the selection. Why are such straightforward
visualization methods preferred to more semantically rich approaches?
Informal discussions with several non-expert users reveal that the
main qualities of visualization applications should be simplicity
of interpretation and manipulation. Objects should be easy to identify
with their attributes and links should be avoided because they require
an effort of concentration. Moreover only contextualised useful information
should be displayed. Consequently it is not surprising that all technically
sophisticated method, whatever their scientific interest, fall short
of being popular in most applications. They could be preferred to
traditional lists or grids if 1) their complexity was limited and
2) they could provide new services that balance some still necessary
efforts of interpretation on behalf of users. 

In this paper we present a new Galois lattice visualization method
which tackles this double challenge. For the sake of simplicity it
turns Hasse diagrams into object grids without loss of expressivity.
The objective is to enrich the popular grid approach with the Hasse
diagram power of expression. We still display Galois lattices as Hasse
diagrams, but objects (not concepts) are visible and links are not
shown to users. Moreover, this approach provides new interesting services
which may enhance its interest for users: it is possible to index
objects with objects, and it is possible to spot structures that may
be of utmost interest for some applications such as team organisation
or document diffusion.

In (\citet{10.1109/TVCG.2009.201}) we already introduced a first
version of this visualization method which showed Hasse diagrams as
layered grid displays. The goal was to index objects with other already
indexed objects which were displayed on the Hasse diagram. But we
still made use of links and the display was not contextualized, i.e.
the Hasse diagram was incrementally built with all indexed objects
each time a new object or group of objects was indexed using other
objects. As a result our approach had two limits with regard to users\textquoteright{}
expectations. Links are still a hurdle for interpretation and since
all objects were displayed on the Hasse diagram, scalability was questionable.
Our method, like traditional Hasse diagrams, faced the unavoidable
problem of complexity and scalability. But it had the quality of providing
a good support for fast indexing. The method we present in this paper
introduces important improvements which overcome the two problems
described above. Contextualisation is obtained through the presence
of a virtual probe which represents the user\textquoteright{}s intention.
It is a visual object which by its own presence extracts and organises
a subset of objects according to their attributes. It is also possible
to load the probe with an object to extract similar objects which
are displayed according to a grid based Hasse diagram without links. 

The idea of using virtual probes or magnets for visualizing and/or
retrieving information is not new. (\citet{Miller2004}) introduces
an immersive visualization probe for exploring n-dimensional spaces
when some scalar function is available depending on n variables. In
this solution the probe is not a visible object but a user\textquoteright{}s
3D view point from which it is possible to project the other dimensions
on 3D walls. (\citet{398849}) introduce visual probes for the visualization
of three-dimensional fluid flow fields. In both papers probes are
used to reveal physical phenomena with continuous parameters. In (\citet{Spritzer2008}),
probes are used for extracting sub-graphs from graphs with limited
visual capacities. Magnets which play the same role as probes are
used in (\citet{Yi2005}) to search for multivariate data. Each magnet
represents an attribute (possibly valuated) and two or more magnets
compete on a 2D screen for attracting dots representing multivariate
data. Without being aware of these results we explored such a metaphor
a few years ago with a very similar display for building concept maps
(\citet{Crampes2006}) We then faced a lot of limits among which some
are reported by the authors in (\citet{Yi2005}) such as the expressivity
of the metaphor and the difficulty of interpretation when there are
two magnets, i.e. two attributes. It is worse with three or four magnets
and the display is meaningless beyond four magnets. We then explored
one fixed probe with potentially multiple attributes and Galois lattices
to propose expressive hierarchical displays. This strategy turned
out fruitful for creating dynamic expressive displays. In this paper
we introduce semantic probes in Galois lattices which are complex
semantic structures for experts, to extract Galois sub-hierarchies
with rich semantic and interactive capacities for novices. 

As far as new services are concerned compared to usual list and grids
display, the new method still gives a good support for indexing objects
with objects partly inheriting from the method we presented in (\citet{10.1109/TVCG.2009.201}).
However the indexing is much improved in this version, particularly
as far as scalability is concerned, because it takes advantages of
contextualisation and of the probe\textquoteright{}s presence. As
a second new service which differentiates it from trivial list or
grid displays, it clearly and simply reveals some interesting structures,
particularly in the context of social networks, such as community
detection based on Hasse diagrams (which we just introduced in \citet{25})
and social complementarities which are introduced in the present paper.

As a conclusion to this state of the art, it is worth investigating
recent developments based on faceted data which present some common
features with our approach (\citet{Yee:2003:FMI:642611.642681}) A
set of items is tagged with terms. For example scientific papers are
tagged with their authors and their subjects. Terms are grouped in
orthogonal (i.e. mutually exclusive) subsets called facets in which
they can be selected by users. At the starting point, it is possible
to see the count of all items in each facet, an item being possibly
duplicated in different facets. When selecting a term, the facets
are updated with the remaining items that are tagged with this term.
In FacetMap and FacetLens facets are graphically and dynamically organized
on the screen, each facet occupying an area proportional to its object
count (\citet{Smith2006,Lee:2009:FET:1518701.1518896}). In faceted
data \textquoteleft{}terms\textquoteright{} are equivalent to \textquoteleft{}attributes\textquoteright{}
(or dually \textquoteleft{}objects\textquoteright{}) in Galois lattices
and choosing a subset of terms in different facets is equivalent to
selecting a unique \textquoteleft{}intent\textquoteright{} (or dually
an \textquoteleft{}extent\textquoteright{}) in Galois lattices as
we shall see below. To compare the technologies we will use the faceted
data vocabulary with the words terms and items.

Our approach is different in several respects. First we need not organize
terms (in our case attributes or dually objects) in orthogonal facets;
they may be of any kind and can be organized in a hierarchy only if
it is interesting. Second the screen is mainly occupied in these applications
by facets and not by the items that are searched. Our point of view
is that users should visually focus on what they are looking for and
not the means to get it. Third in FacetMap and FacetLens facets are
graphically represented with bubbles which are dynamically reorganized
when a term is selected. The equivalent in our application is a traditional
hierarchy of terms in alphabetic order because we consider that it
is the traditional and most effective way of finding entities. The
reported evaluations in both papers mention the attractive effect
of the graphical interface, and do not mention usability problems
with reorganisation during experiments. But it is also reported that
some users do prefer lists in alphabetic order to explore terms. We
also observed this users\textquoteright{} expectation and this is
the reason why we present terms in a hierarchical list. However the
main differences with our approach is related to the choice and the
presentation of returned items. The facet approach is a way of presenting
an \textquoteleft{}AND\textquoteright{} choice of terms and the selected
items are displayed in a list. Thanks to the Galois lattice theoretical
basis our probe approach displays selected items in layers with different
levels of match corresponding to all possible combinations of terms
(conjunctions and disjunctions) and not only a unique Boolean conjunction.
The probe display opens up other functional possibilities such as
weighting terms, indexation of items with items, items\textquoteright{}
complementarities, etc. which are difficult or impossible to obtain
with sole conjunctions of terms. Faceted data still remains an interesting
approach. Some new functions have been introduced in recent facet
driven applications (\citet{Lee:2009:FET:1518701.1518896}) such as
linear facets and pivoting. They reinforce the interest in this technology.
But the set conjunction paradigm remains different from our Galois
lattice based paradigm.

\subsection{Concept Hierarchies\textquoteright{} basics }

In Formal Concept Analysis (\citet{GanterB.1999}), a finite set of
\textquoteleft{}objects\textquoteright{} with \textquoteleft{}attributes\textquoteright{}
can be organized in a lattice of \textquoteleft{}concepts\textquoteright{}
that contain these objects according to their attribute commonality.
The objects (respectively the attributes) are called formal insofar
as they may be real objects or abstract objects (respectively attributes).
Many domains are concerned, such as tagged photos, videos or documents,
hospitals and patients, social networks, medical data, etc.

The organization process starts with a formal context, i.e. a table
with the objects as rows and the attributes as columns. Any entry
is marked (e.g. a cross or 1) if the corresponding object possesses
the corresponding attribute, and is not marked (e.g. 0) if the object
does not possess such an entry. Formally, a formal context is a triple
$(G,M,I)$ where G is a set of objects, M a set of attributes and
I is a binary relation between the objects and the attributes, i.e.
$I\subseteq G\times M$. Table \ref{tableau1} presents a formal context
taken from a toy example where the set of objects G is a set of 4
actors, the set of attributes M is a set of 6 films, and the relation
$(g_{i},m_{j})$ is valued 1 if the actor $g_{i}$ played in the film
$m_{j}$, and 0 otherwise. We give the same name I to the binary relation
and the incidence matrix it defines. 

\begin{table}[h]
\centering{}%
\begin{tabular}{|l||l||l||l||l||l||l|}
\hline 
 &
Film1 &
Film2 &
Film3 &
Film4 &
Film5 &
Film6\tabularnewline
\hline 
\hline 
Brad &
1 &
1 &
1 &
0 &
1 &
0\tabularnewline
\hline 
\hline 
Angelina &
1 &
0 &
1 &
0 &
1 &
0\tabularnewline
\hline 
\hline 
Cate &
1 &
0 &
0 &
1 &
0 &
0\tabularnewline
\hline 
\hline 
Leonardo &
0 &
1 &
0 &
1 &
1 &
1\tabularnewline
\hline 
\end{tabular}\caption{\label{tableau1}A small context with films and actors}
\end{table}

The next step in building the concept lattice is to define concepts
according to \citet{GanterB.1999}. A concept is a pair of subsets:
a subset of objects O$_{i}$ (called the extent) and a subset of attributes
A$_{i}$ (the intent) that the objects share. Two operators both denoted
by $\lyxmathsym{\textquoteright}$ connect the power set of objects
$2^{G}$ and the power set of attributes $2^{M}$:

$\lyxmathsym{\textquoteright}:$$2^{G}\rightarrow2^{M},\: O_{i}\lyxmathsym{\textquoteright}=A_{i}={\{m\in}$M${|\forall g\in}$
O${_{i\:},g}$I${m\}}$

Dually on attributes:

$\lyxmathsym{\textquoteright}:$$2^{M}\rightarrow2^{G},\: A_{i}\lyxmathsym{\textquoteright}=O_{i}={\{g\in}$G${|\forall m\in}$
A${_{i\:},g}$I${m\}}$

Informally applying the operator \textquoteright{} to a subset O$_{i}$
of objects of G extracts the subset A$_{i}$ of attributes of M that
are shared between all objects of O$_{i}$ and conversely A$_{i}\lyxmathsym{\textquoteright}$
identifies all objects (the subset O$_{i}$ of G) who share the same
subset of attributes A$_{i}$ of M. The composition operators \textquoteright{}\textquoteright{}
are closure operators (idempotent, extensive, and monotonous), which
means that A$_{i}\lyxmathsym{\textquoteright\textquoteright}=$A$_{i}$
and O$_{i}\lyxmathsym{\textquoteright\textquoteright}=$O$_{i}$ for
any (O$_{i},$A$_{i})\subseteq$ G$\times$M. These operators \textquoteright{}
will be important for the properties of our model below. 

For the context presented in table \ref{tableau1}, the concepts are
nodes in the line diagram shown in figure \ref{figure2}, such as: 

$Concept6=(\{{Angelin,Brad}\},\{{Film1,Film3,Film5}\})$

where $\{{Film1,Film3,Film5\}}$ is the intent and $\{{Angelina,Brad}\}$
is the extent: 

${\{Film1,Film3,Film5}\}\lyxmathsym{\textquoteright}=\{{Angelina,Brad}\}$

$\{{Film1,Film3,Film5\}}\lyxmathsym{\textquoteright\textquoteright}={\{Angelin,Brad}\}\lyxmathsym{\textquoteright}={\{Film1,Film3,Film5\}}$

Following the process of concept identification, the next goal is
to build a lattice whose elements are the concepts. A partial order
on formal concepts is defined as follows: 

(O$_{2},$A$_{2})\le($O$_{1},$A$_{1})\: iff$ O$_{2}\subseteq$
O$_{1}$ (and consequently A$_{1}\subseteq$ A$_{2}$). 

The ordered concepts form a complete lattice called a \char`\"{}concept
lattice\char`\"{}. Figure \ref{figure2} shows the concept lattice
of the toy film-actor example as a Hasse diagram. The set of concepts
L is completed if necessary by a top concept that contains all objects
and a bottom concept that contains all attributes. A Hasse diagram
is a graph whose vertices are the concepts, ordered from top to bottom
according to their order in the lattice; the edges are drawn between
concepts when two concepts are directly ordered without transition
through another concept. In our example each concept is a group of
films with their common actors. As can be seen in figure \ref{figure2},
an object (an actor) as well as an attribute (a film) may appear in
several concepts. 

The transpose of the context matrix produces a Galois Lattice which
is the dual of the original context. The roles of objects and attributes
are reversed. The Hasse diagram\textquoteright{}s structure is the
same; it is just turned upside-down. In our example if we place concept-9
at the top, concept-0 at the bottom and accordingly reorganise the
Hasse diagram, films become objects and actors become attributes because
tradition applies the object order for the top down hierarchy. This
is a slight problem for our presentation. After experimenting with
users it appeared that the probe we will introduce below should be
placed at the top of the screen. We will see below that when searching
for films the probe must be loaded with actors. Consequently, since
we want to be sound with Formal Concept Analysis in the present paper,
actors must be defined as objects and films as attributes, although
the search applies to films. We will choose objects to search attributes.
This is purely formal because attributes and objects play dual roles
and final users are not concerned by this vocabulary; they only know
their domain of application vocabulary, such as films/actors, people/competences,
papers/authors, etc. 

\begin{center}
\begin{figure}[h]
\centering{}\includegraphics[width=8cm]{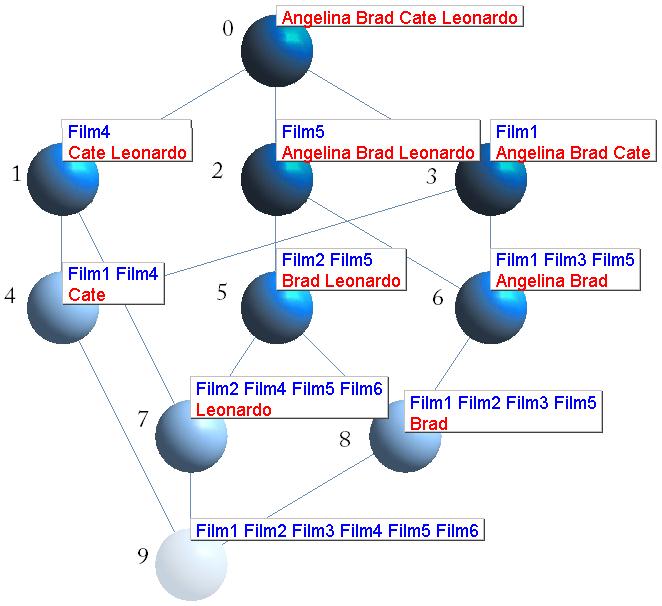}\caption{\label{figure2}A line diagram with concepts showing intents and extents}
\end{figure}

\par\end{center}

\subsection{Visualizing Galois lattices}

Many methods have been proposed for building line diagrams representing
Galois lattices such as incremental building (\citet{Godin95incrementalstructuring})
or Force Directed Placement (\citet{FREESE,Hannan:2006:SLD:2180241.2180261,Kamada:1989:ADG:63827.63830}).
Their main goals are algorithmic efficiency and display quality (see
(\citet{Arevalo_sigayreta.:}) and (\citet{Kuznetsov2002}) for a
survey of some algorithms and their performances). Although Galois
lattices are mostly targeted to objects with Boolean attributes, they
may also be used for organizing multi-valued data (\citet{GanterB.1999})
and even hybrid data (\citet{Villerd2009}). Several tools implement
these methods among which we used two of the most widely known in
the FCA community for illustrating examples in this paper. We took
advantage of Lattice Miner (\citet{Lahcen2010}) because it is recent
and, beyond a real aesthetic effort, it proposes many visualization
options that we use for better illustrations of simple examples. However
the number of concepts it can compute is limited. Consequently we
also used Galicia (\citet{Valtchev03galicia:an}), to create the Hasse
diagrams in Figure \ref{figure1}. 

Even with small examples it is not obvious for non-experts to analyse
the Galois lattice conceptual structure and navigation is not easy
when looking for particular sets of objects or attributes. However
it is possible in Galicia or in Lattice Miner to interact with a concept
and display its intent and extent but at the expense of other problems:
edges are hidden and information is getting cluttered even in little
Galois lattices. Some authors have explored better design and interaction
in other FCA environments for helping non-experts browsing Galois
lattice such as in \citet{eklund04information}. Although these authors
report positive results, Galois lattices which are tested are small
and scalability remains an open question. 

Real applications require bigger contexts. To better experiment with
scalability we built a benchmark with a medium context containing
127 films (attributes) and 245 actors or directors (objects). The
resulting Galois lattice built with Galicia is presented in figure
\ref{figure1}. The nice diamond shape with three intermediate layers
is exceptional. It reflects the fact that we we considered two actors
and one director for each film all films with two actors and a director,
except for the three films of the Ocean\textquoteright{}s series for
which we considered five actors (two concepts concerning these films
are visible on a small fourth layer). Real applications present more
complex Galois lattices with no particular symmetry. We built such
a simplified benchmark structure for the following reason. Visual
analysis is difficult on Galois lattices and their Hasse diagram display
using tools like Galicia or Lattice Miner. Conversely the semantic
probe is not affected by this problem. As a result in order to build
experiments with users and challenge our semantic probe on traditional
Hasse diagrams we had to build a simplified benchmark to the detriment
of the probe. If experiments give better results on such simplified
data with the probe, it would also be the case for more general data.

As far as navigation is concerned, Galois lattices\textquoteright{}
scalability is even worse. To overcome this problem several approaches
have been proposed such as focus \& context and fisheye in Lattice
Miner (\citet{Lahcen2010}). Only experts can however analyse the
resulting display. Other navigation applications which are targeted
to novices propose a local concept approach. A user\textquoteright{}s
query is considered as a set of attributes. The corresponding extent
is displayed with facilities for removing attributes or adding attributes
from the list of descendant concepts. 

As a result the user can navigate upward and downward on the Galois
lattice without ever seeing it such as in the experiments conducted
in \citet{Godin:1993:ECN:172206.172208}, the Credo application (\citet{Carpineto2004})
or in the more recent application ImageSleuth (\citet{Ducrou:2006:IUI:1793623.1793625,DBLP:conf/iccs/DucrouVE06}).
But user\textquoteright{}s navigation is entirely limited to one concept
at a time, and all conceptual structures have disappeared. 

Coming back to a global Galois lattice view, several reduction algorithms
have been described in the literature for managing scalability. Four
of them are frequently applied. They are introduced in the next section.

\subsection{Galois lattice reduction and other methods \label{section23}}

\subsubsection{Nested, iceberg and stability based reductions}

Nested line diagrams are constructed when it is possible to extract
sub-contexts and partition the attribute set (\citet{GanterB.1999}).
Resulting line diagrams are clearer but to the detriment of easiness
of navigation and understanding. Iceberg lattices reduce Galois lattices
to a subset of concepts whose intent\textquoteright{}s support count
is above a user defined threshold (\citet{Stumme2002}). The support
count of an attribute set $A_{i}$ is define as: $support(A_{i})=|A_{i}\lyxmathsym{\textquoteright}|/|G|$.
A concept is frequent if its intent is frequent, i.e. there are many
objects with the corresponding set of attributes compared to other
concepts. The set of frequent concepts of a context is called the
iceberg concept lattice of the context. Reducing a Galois lattice
to an iceberg lattice is efficient when looking for association rules
to the expense of missing rare information. Another reduction process
is based upon stability whose definition is formally less intuitive
as support (\citet{Kuznetsov2007,roth2006towards}). Intuitively,
a concept is stable inasmuch its intent is found in many combinations
of objects from its extent. This reduction process is particularly
interesting for data and knowledge mining (\citet{Jay2008}), but
its visual efficiency is limited, depending on a user defined threshold,
and it loses rare information which may be highly interesting in many
applications.

\subsubsection{Object or Attribute Galois sub-hierarchies \label{section232}}

Extracting Object or Attribute Sub-Hierarchies is another reduction
process for pruning Galois lattices. It presents a remarkable advantage:
contrary to iceberg or stability driven reduction, there is no loss
of information (\citet{Godin95incrementalstructuring}). 

The reduction process is based upon the observation that many objects
and attributes belong to several concepts. In our example attribute
Film5 for instance belongs to concepts 2, 5, 6, 7, 8 and 9. It is
possible to get rid of this redundancy without loosing information.
The reduced intent (respectively extent) of a concept$(O_{i},A_{i})$
is the set of objects (respectively attributes) that belong to A$_{i}$
(respectively O$_{i}$) and do not belong to any upper (respectively
lower) concept. In the following we will only consider attribute reduction,
the same results being dually possible with objects. In the example,
the reduced intent of concept6 is \{Film3\} since Film1 and Film5
belong to lower concepts (respectively concept3 and concept2).

\begin{figure}[h]
\centering{}\includegraphics[width=8cm]{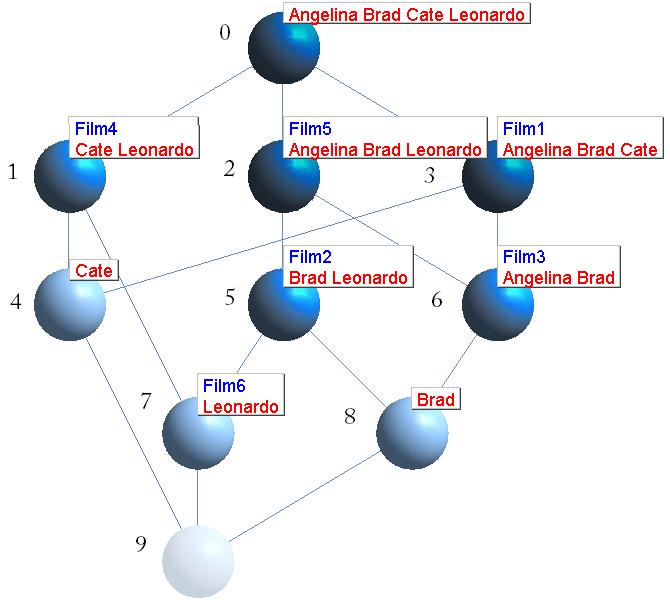}\caption{\label{figure3}An Attribute Galois Sub-Hierarchy}
\end{figure}

For each attribute (a film in our example), there exists a unique
attribute-concept that represents the most specialized concept that
contains the attribute. Figure \ref{figure3} shows the Galois sub-hierarchy
which is derived from the Galois lattice in figure \ref{figure2}.
Films appear in only one concept although they are implicitly present
in other concepts. Since we want attributes to only appear once, only
concepts with attributes are kept in the lattice and other concepts
can be rebuilt through inheritance. In our example, concept4 which
was originally $({\{Cate\}},{\{Film1,Film4\}})$ is now $(\{{Cate\}},\{{}\})$
with an empty reduced intent, and its original intent ${\{Film1,Film4\}}$
can be rebuilt through the union of concept3 and concept1\textquoteright{}s
intents. This act of pruning when applied to attributes or objects
is the one proposed in \citet{Godin95incrementalstructuring} under
the name PCL/X. The new line diagram is a particular case of what
is called a Galois sub-hierarchy (\citet{Godin:1993:BMA:167962.165931}).
It is a lighter visualization of data when only focusing upon the
attributes (dually the objects), in our case the films (dually the
actors and directors). To our best knowledge, attribute or object
driven reduction process has only been applied for building incremental
Galois lattices. In \citet{10.1109/TVCG.2009.201} we used it to organize,
visualize and index social photos. However the display was not user
centered and scalability was a remaining issue which we address in
this paper. 

Figure \ref{figure3} is clearer with no redundancy on films. Thanks
to the edges one can see for example that Leonardo played in Film6,
Film2, Film5, and Film4. But catching this knowledge is not immediate.
Moreover in a realistic context edges would be covered by concept-nodes.
In that case a good thing would be to get rid of edges without losing
the possibility of identifying concepts. This is what we are going
to do with semantic probes that we present in the following section.

\section{SEMANTIC PROBES }

The semantic probe model and techniques introduce a user centric approach
of Galois lattices which is easy to understand for novices because
it is not concept oriented and it has no edges. The display clearly
shows entities that are searched without repetition and without the
necessity of following edges.

Let G be a set of objects and M a set of attributes, each object being
characterized by a subset of these attributes. Objects and attributes
are represented by words or icons depending on the application domain.
We define a semantic probe P as a bag which is loaded with some objects
representing a particular focus of interest for a user and with which
it is possible to interact. The corresponding objects' attributes
react and gather around the probe as if it were a magnet. Remember
that the terms 'objects' and 'attributes' are formal and the roles
are dual. We chose in this description to load the probe with objects
and to attract attributes to comply with the FCA tradition which places
all objects at the top of the hierarchy. If we had placed the probe
at the bottom, we would have loaded it with attributes and have attracted
objects. This last observation will be of great interest at the end
of the paper.

Formally, we define a semantic probe P as follows: 

Let (G,M,I) be a formal context with G a set of objects, M a set of
attributes, and I$\subseteq$G$\times$M an incidence relation, 

A probe P is a bag which, when loaded with a set of objects $G={\{g_{i}\}},G\subseteq$G,
produces two results: a sub-context and a Galois sub-hierarchy display.

\subsection{The sub-context }

The probe P\textquoteright{}s set of attributes defines a new context
$(G,M,I)$ which is a sub-context of the original context (G,M,I)
where: 
\begin{itemize}
\item $G$ is P\textquoteright{}s set of objects, 
\item $M={m_{j}}$ is a subset of M: $M\subseteq$M, $m_{j}\lyxmathsym{\textquoteright}\cap$G$\neq\textrm{Ø}$ 
\item $I$ is an incidence matrix whose rows are the rows of I corresponding
to the objects belonging to $G$ and whose columns are the columns
of I corresponding to the attributes belonging to $M$. From this
sub-context it is possible to build a Galois lattice G$_{p}$ and
create an original layered display.
\end{itemize}

\subsection{Semantic Probe\textquoteright{}s object-concept display }

\begin{figure}[h]

\centering{}\includegraphics[width=10cm]{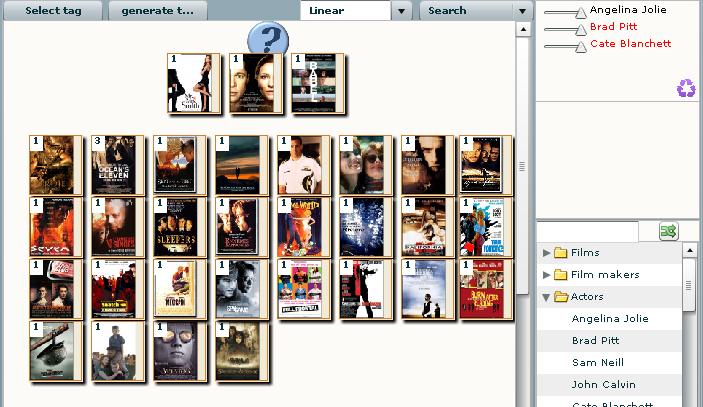}\caption{\label{figure4}A semantic probe display}
\end{figure}

In figure \ref{figure4} the general context is the whole benchmark
containing 127 attributes (films) and 245 objects (actors or directors).
The probe which is represented by the blue button with a question
mark at the top is loaded with the object subset $G={\{AngelinaJolie,BradPitt,CateBlanchett\}}$
which defines a sub-context. All attribute-concepts whose extent contains
one or more selected objects slide up. Each attribute-concept is a
group of attributes (DVD jackets) which share exactly the same objects
(actors and directors) in the original context. For the sake of communication
with lay users we use the word 'group' rather than 'attribute-concept'
or 'concept'. A group is represented by the jacket of one of its DVDs.
The figure at the top left of the group's picture indicates the number
of DVDs it contains. In this particular benchmark which was created
for experimentation all film castings are different but for the three
films from the Ocean\textquoteright{}s trilogy. Consequently all group
pictures but one display the number 1 and the Ocean\textquoteright{}s
trilogy displays the number 3. 

\begin{figure}[h]

\centering{}\includegraphics[width=10cm]{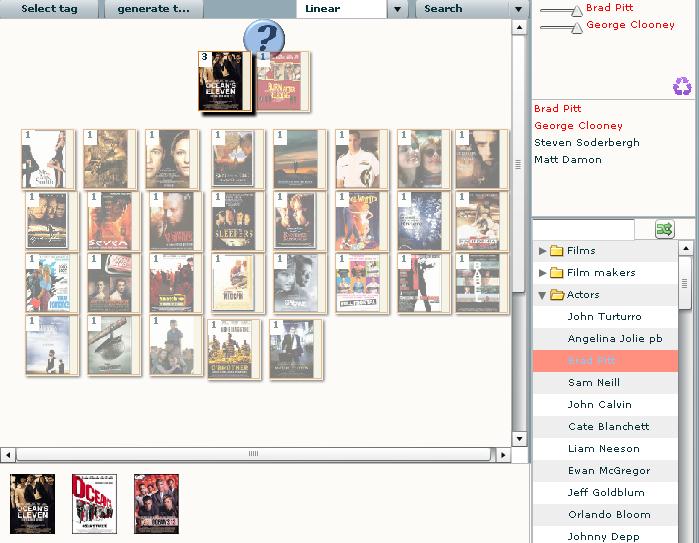}\caption{\label{figure5}Clicking an object-concept reveals its intent and
extent}
\end{figure}

When clicking a group, a pane opens up at the bottom. It displays
the group\textquoteright{}s DVDs. Since a group represents an attribute-concept
from the whole context, clicking actually reveals its intent at the
bottom and its extent in the middle right pane. Figure \ref{figure5}
shows the attribute-concept whose intent is the Ocean\textquoteright{}s
trilogy shown in the bottom pane and the extent is a set of 4 people
shown in the middle right pane. Two characters are red. They are those
that are included in the probe whose extent is shown in the upper
right panel. As a result comparing the two upper right panes it is
possible to identify the objects that are common to the group and
the probe (red), the objects that are in the probe and are not in
the group (black in the upper right pane in figure \ref{figure6})
and the objects that are in the group and not in the probe (black
in the middle right pane in figure \ref{figure6}).

The core of the display strategy is to place the groups at a distance
from the probe according to their semantics (the extent) and the probe\textquoteright{}s
semantics. Let ${a_{i}}$ be a group A\textquoteright{}s extent. We
define the Semantic Distance between the probe and the group as follows: 
\begin{defn}
\textbf{$SD(P,A)=\frac{1-\left|G\cap\{a_{i}\}\right|}{\left|G\right|}$}

where $G$ is the probe\textquoteright{}s extent.
\end{defn}
All groups which are at the same semantic distance are put in a common
layer. All layers are placed from top to bottom according to their
semantic distance, the layer at the top being the one with the smallest
distance to the probe. All groups belonging to a layer are then placed
in a grid which clearly identifies them. In Figure \ref{figure4}
three groups are visible in the first layer, and 25 groups in layer
2. The probe displays well identified entities, in this example DVDs,
when the traditional Galois hierarchies display concepts with no easy
means for novices to identify objects or attributes. 

The probe is equivalent to the top concept of a Galois lattice as
in figure \ref{figure3}: it contains the set of objects from which
the hierarchy is built. The ordering from top to bottom is linked
to a decreasing number of objects. The result is a balance between
the search engines\textquoteright{} traditional display and the rich
conceptual display of Galois lattices. The core idea is to invite
users to interact with Galois lattices as if they were interacting
with traditional displays.

\subsection{Probe\textquoteright{}s concept visualization through interactions }

Suppose we want to know in which films a particular actor, say Brad
Pitt, played. Each group is an original concept with the same subset
of attributes and the same subset of objects. All groups have different
extents and intents. To search for films in which Pitt\textquoteright{}s
acted it is possible to drag the object \textquoteleft{}Brad Pitt\textquoteright{}
to the empty probe to get all groups containing films with at least
Brad in the extent. Consequently Brad may be in several groups of
films with other actors. 

\begin{figure}[h]

\centering{}\includegraphics[width=10cm]{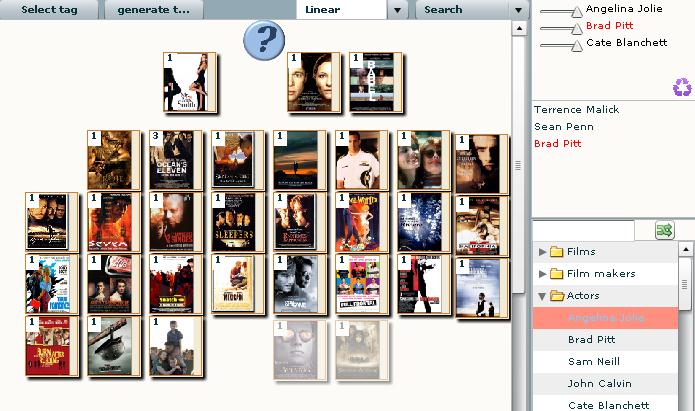}\caption{\label{figure6}Interaction reveals probe centred concepts}
\end{figure}

But when the probe is already loaded with several objects like in
figure \ref{figure6} there are groups in the lower layers which contain
other objects than the interesting one. Therefore we are only interested
in a subset of the visible groups. To reveal this subset we use the
fact that the display is a sub Galois lattice with attribute-concepts,
the whole objects set being the probe\textquoteright{}s content: we
are looking for the intent of a scattered concept whose extent is
Brad.

As it was explained in section \ref{section232}, an attribute sub-hierarchy
shows intents without attribute redundancy and the concepts\textquoteright{}
intents of this sub-hierarchy are only visible through inheritance.
The concepts\textquoteright{} intents we are looking for when searching
DVDs with \textquoteleft{}Brad Pitt\textquoteright{} may be scattered
within layers and between layers. To manage this difficulty we apply
the following design strategies. First all groups in the same layer
belonging to the same probe-filtered concept, i.e. having the same
subset of objects common with the probe, are dynamically regrouped
side by side (this dynamic regrouping is very spectacular and well
appreciated by users). They are optionally separated by blank objects
from other groups when probe-filtered extents are different. Second,
we showed in section \ref{section23} that a concept can be rebuilt
from an attribute-concept through the union of its parents in the
hierarchy. Practically, when the user hovers with the mouse over a
group in the probe\textquoteright{}s induced hierarchy it is possible
to reveal the corresponding probe centered concept to which this group
belongs through the following visual effects. The common objects with
the probe\textquoteright{}s objects are turned into red in the two
upper right panes to show that they are the probe\textquoteright{}s
driven concept\textquoteright{}s extent. All other groups which do
not belong to the concept are partly turned transparent. The remaining
clearly visible groups define the concept\textquoteright{}s intent
whose extent is the intersection between the selected group\textquoteright{}s
extent and the probe\textquoteright{}s extent. Figure \ref{figure6}
shows the concept extracted from the probe-driven sub-hierarchy when
the user hovers with the mouse over the group represented by the film
\textquoteleft{}Seven\textquoteright{}. The only common object between
this film and the probe\textquoteright{}s extent is $\{{Brad\: Pitt}\}$.
The corresponding probe-driven extent ${\{Brad\: Pitt\}}$ is highlighted
in red in both side panes. The concept intent reveals three groups
in the upper layer, and 26 in the lower layer. All these groups and
the probe have ${\{Brad\: Pitt\}}$ as a common set of objects. If
the user hovers over \textquoteleft{}Babel\textquoteright{} in the
top layer, only the groups represented by 'Benjamin Button' and 'Babel'
will appear. 'Brad Pitt' and 'Cate Blanchett' will be highlighted
in red. The corresponding concept is (${\{BenjaminButton,Babel\}}$,
$\{{BradPitt,CateBlanchett}\}$). 

This interaction for revealing a concept is attribute-driven since
it is necessary to hover with the mouse over a group. It is also possible
to apply an extent oriented way of revealing a concept. The probe\textquoteright{}s
objects in the top right probe pane are endowed with sliders (see
Figures). Dragging a slider to 0 turns the corresponding object down
and all the groups with this object slide down. The remaining groups
by the probe define the intent which corresponds to the probe\textquoteright{}s
extent whose objects\textquoteright{} \textcolor{black}{weight} is
equal to 1. The sliders can also be set to a value between 0 and 1.
Groups on the same layers are separated into those that do not contain
the modified value which stay at their level, and those that slide
down but are still on the screen. This advanced interaction was activated
by the user searching for personnal Facebook photos in figure \ref{figure15}.
This is also what is applied to separate Arsenal from Manchester in
the industrial prototype presented in figure \ref{figure16}. It must
be noticed that to our best knowledge this method of weighting objects
in concepts\textquoteright{} extents (or dually in concepts\textquoteright{}
intents) to reveal sub-structures in Galois lattices is new even in
the Formal Concept Analysis community. Our approach provides a new
way of seeing and weighting Hasse diagrams. Moreover this reorganization
provokes an impressive animation on the screen which is very appreciated
by users. 

The interaction and visual effects described above, which reveal concepts\textquoteright{}
intents avoid edges\textquoteright{} visual complexity. Their drawback
is that even if they are simple, their interpretation is not so obvious.
We do not yet know to which point it is interesting to give these
conceptual clues. However several presentations and experiments with
users have shown that it is better to use sliders than transparency
to reveal concepts. This is an important point for deploying the technology.

\subsection{Interactions and navigation }

\begin{figure}[h]

\begin{centering}
\includegraphics[width=10cm]{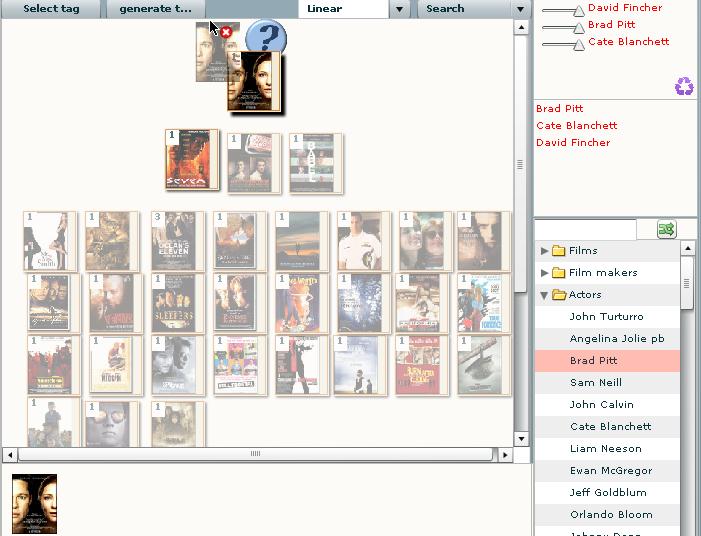}\caption{\label{fig7}Loading the probe with objects}

\par\end{centering}

\end{figure}

Interacting with traditional Galois lattices is seldom mentioned in
the literature although some applications like Lattice Miner offers
a few limited possibilities. The probe driven display with explicit
intents are not only simple and easy to understand compared to traditional
Galois lattices. They are also particularly useful for interacting
with all objects and attributes. Users can change a probe\textquoteright{}s
semantic state through different interactions: 
\begin{enumerate}
\item Adding an object to the probe\textquoteright{}s semantics by double
clicking onto it in the tree of objects at the bottom right, or, after
clicking onto a group, selecting an object from the group\textquoteright{}s
object list in the central right panel, then dragging and dropping
the new object onto the probe. This second possibility is particularly
interesting because a group may suggest new objects for searching
other groups.
\item Removing an object from the probe through dragging and dropping it
onto the bin in the probe\textquoteright{}s object pane. Double clicking
onto this bin removes all objects from the probe.
\item Adding a group\textquoteright{}s extent to the probe through dragging
and dropping the group\textquoteright{}s image from the hierarchy.
As a result all the group\textquoteright{}s objects which were not
already part of the probe\textquoteright{}s semantics are added to
it (see figure \ref{fig7}). This last interaction is original, particularly
useful and well understood by testers and users.
\item Weighting tags in the probe's extent for separating groups in the
same layer.
\end{enumerate}
Updating the sub-hierarchy is made after the end of the interaction.
If groups must disappear because they have no common objects with
the probe, they slide down and hide. If new groups are eligible, they
slide up and find their proper place in the hierarchy. Other groups
may change smoothly of place in the hierarchy, changing of layer or
creating a new layer. All movements are made of fluid aesthetic animation
to maintain the user\textquoteright{}s mental map. These animations
are particularly appreciated both during presentations and tests with
users. 

\subsection{Semantic probe\textquoteright{}s qualities and novelty }

Our goal was to define an environment whereby Galois lattices, which
are sophisticated experts\textquoteright{} tools, are simply used
by lay users. The probe\textquoteright{}s metaphor and display show
the usability qualities which are expected by them as explained in
section\ref{section2}: 

\textbf{Contextualization}: Only attributes (DVD jackets in the example)
that meet totally or partially the probe\textquoteright{}s semantic
profile (actors and directors) are displayed. 

\textbf{Reification}: It is possible to easily identify attributes
or groups of similar attributes with their objects and without redundancy. 

\textbf{No edges}: Contrary to usual Hasse diagrams and the solution
we proposed in \citet{10.1109/TVCG.2009.201} there are no visible
edges. Edges are difficult to read and understand for lay users. In
our application they are replaced by the probe\textquoteright{}s profile
combined with the navigation tools. 

These simplification improvements are achieved with little loss of
conceptual information which distinguishes our approach from a trivial
list or grid:

\textbf{Conceptual structure}: Concepts and concept relations are
revealed through the regrouping of attributes and interactions as
shown in section 3, or through the use of sliders. 

\textbf{Navigation}: The display gives conceptual hints and provides
interaction capacities for facilitating navigation when placing objects
on to the probe.

\textbf{Mental map}: The soft animation maintains the user\textquoteright{}s
mental map when groups are reorganized after a modification of the
probe\textquoteright{}s profile. This feature is particularly attractive
when shown during presentations. Its interest is not only limited
to aesthetic animation. Contrary to a list or a grid presentation
of results after a query in traditional search engines, it only shows
what is changed in the results and how these changes occur.

All these qualities show first that a probe driven sub Galois lattice
display without edges meets most simplicity criteria that lay users
are looking for, and second that it provides a better approach for
navigation in an object-attribute database. Next section introduces
several tests and industrial experiments that have been conducted
for verifying the above hypothesis.

\section{Tests and applications}

Our first goal for conducting tests was to compare the probe paradigm
with its two main competitors: Galois lattice based navigation and
traditional Boolean querying using index terms, the last one being
the most widespread mode of searching databases when indexes are available. 

As far as other technologies are concerned such as faceted data, the
goal was not to check whether the probe approach is more efficient
or more attractive, though some experimental results with these technologies
are worth being mentioned. For instance two faceted data applications
are compared in \citet{Smith2006} using a group of 10 participants
well aware of computer interfaces. Memex is a text oriented faceted
data browser whereas FacetMap presents adaptive bubbles representing
facets on the screen. Results do not reach significant conclusion
about the success of a particular technology. But authors are more
interested in the formative results given by the testers\textquoteright{}
observations. Other formative experiments are conducted in \citet{Lee:2009:FET:1518701.1518896}
with FacetLens, which extends FacetMap. Six people are involved in
the test, none of them lay users. Reported usability results are interesting,
but no comparison is made with other applications, such as with traditional
Boolean search engines. 

Focussing on our Galois lattices (GLs) experimental context we mostly
find experiments on local navigation around concepts. In \citet{Godin:1993:ECN:172206.172208}
local navigation on GLs is compared to two more conventional retrieval
methods: hierarchical classification retrieval and Boolean querying
with index terms. Their result show that local navigation on GLs outperforms
hierarchical classification navigation, but it does not do better
compared to Boolean querying. A more recent experiment in \citet{Ducrou:2006:IUI:1793623.1793625}
is conducted with the ImageSleuth application involving 29 testers.
GL based local navigation is compared to hierarchical classification
navigation. Authors provide similar results: local navigation on a
GL gives better results than hierarchical classification navigation.
No comparison is given with Boolean querying with index terms when
according to \citet{Godin:1993:ECN:172206.172208} this approach is
more efficient than hierarchical classification navigation. In our
case navigation is performed through extracting a sub hierarchy and
organizing it under a probe; it is in between a Hasse diagram search
which represents a Galois lattice and local search on individual concepts.
Taking into account all these experiments, the conclusion is that
Boolean querying is the search method to challenge because it is not
clearly outperformed by any of these technologies and it is still
the most widespread. However since the probe rebuilds parts of a Hasse
diagrams without lines between concepts, it is also necessary to compare
it with traditional Hasse diagrams. 

Two phases of tests were conducted. The first phase did not require
new developments beyond the prototype and could be organized within
the laboratory. The second phase required new developments and the
support of an industrial. 

The first phase targeted two questions: 
\begin{itemize}
\item 1) Is it possible for lay users to navigate on a Galois lattice when
using our semantic probe compared to traditional Hasse diagrams. 
\item 2) Does the probe approach equal Boolean search with index terms for
traditional tasks and does it outperform it for some tasks. 
\end{itemize}
The second phase had more open goals: 
\begin{itemize}
\item 3) Application on real data: is the probe interesting for users using
their personal data? 
\item 4) Deployment: for what sort of applications can the probe approach
be the most efficient? 
\item 5) New services: is it possible to imagine new services for which
traditional Boolean search engines are not or are poorly adapted? 
\end{itemize}

\subsection{First phase}

\textbf{Methodology }

\medskip{}
The first phase required testing our probe environment against navigation
on a Hasse diagram, and then against a Boolean search engine. Twelve
students studying general engineering ageing from 20 to 23 (including
4 females), were asked to answer questions from the database of 127
films and 245 actors or directors with the support of the three technologies.

The first method consisted in navigating on a whole Hasse diagram
of the film database. As it was already mentioned in the paper the
database had been built for helping users navigating on such a structure
which may be very complex even on limited contexts. The concept hierarchy
is very symmetric and there are few layers (see Figure 1).We used
Galicia for building this Hasse diagram. Quickly it appeared during
the tutoring preparation that explaining what the line diagram meant
and how to navigate took a long time. Moreover none of them could
properly navigate on the Hasse diagram. We tried using Lattice Miner
which proposes advanced filtering and navigation tools. The tool could
not build the lattice because there were too many concepts on standard
PCs. The test required that the computer had to be of the kind used
by lay users and usability study on powerful computers was out of
question. 

In conclusion, although \citet{eklund04information} suggest that
navigation on very simple Hasse diagrams is possible for lay users
with the hypothesis that scalability should not be a problem, our
tests show a different result. Navigation on medium size Hasse diagram
is complicated. We now focus on the second question which assumes
that it is possible to search with the probe. 

To answer the second question we had to compare navigation using the
probe with a Boolean engine. We chose Amazon because it is widely
used. There may be more efficient search engines, but our purpose
was not performing a general test. We only needed a well known and
widely used engine. Moreover the film database which is used in our
examples and which is used for testing had been built in the first
place using Amazon. We knew that there would not be biases from the
data.

Each tester had to answer a set of questions using both environments.
The subjects were asked not to tell other testers what was taking
place and what questions were asked. The subjects had to draw lots
for the order of the environment to assess to avoid any possible biases.
The response time for Amazon was also tested prior to commencement
to ensure that the two applications were comparable in terms of response.
None of the subjects had previously been exposed to the probe application
prior to assessment and only a few used Amazon. Consequently, the
test started with an explanation read to each subject individually
and a short demo on the two applications was provided, even for those
who had already had experience with Amazon. 

After each test some measures were taken, such as the time for obtaining
the answer and the quality of the answer (number of mistakes or failure
to give an answer after a time delay). We applied a delay of one minute
or two minutes for answering to mimic the fact that lay users are
known for abandoning a tool if the service is not quickly given either
because the application is too complicated or because they have difficulties
to use it. They were also asked to assess the degree of confidence
they gave to their answers. At the end of the test, some qualitative
questions were asked, comparing the two methods. 

We applied ``repeated measures t-tests'' with unequal variances
to the results of the tests for each questions when enough paired
values were available (results show that it is not always the case
due to the methodology of giving a time delay for the answers). H0,
the null hypothesis, asserts that the difference between two responses
measured on the same statistical unit has a mean value of zero. 

\medskip{}
\textbf{Results}

\medskip{}
\textit{Q1: \textquotedblleft{}Cite two films in which Ben Stiller
played\textquotedblright{} }

The real objective of this simple first question was to train the
subjects on both environments. All subjects managed to give an answer
with a mean time of 18.7 seconds for Amazon and 16.9 seconds for the
probe. The mean times\textquoteright{} difference is not significant
(confidence in H0: p = 0.13).

After this first question testers were also invited to freely explore
the data with other actors and films to get used to the environments.
They could do it without any problem on both environments. Consequently
as far as the probe is concerned we could conclude that it is possible
to navigate on a Galois lattice with the probe when it is difficult
and even impossible with the whole Hasse diagram on a medium size
database. 

The three following questions were of increasing complexity: 

\begin{figure}[h]
\centering{}\includegraphics[width=4.2cm]{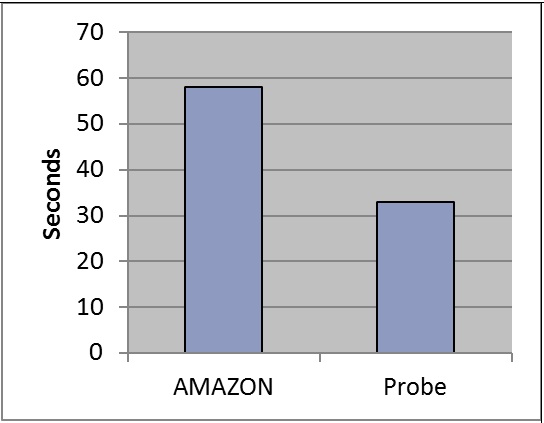}\includegraphics[width=5.2cm]{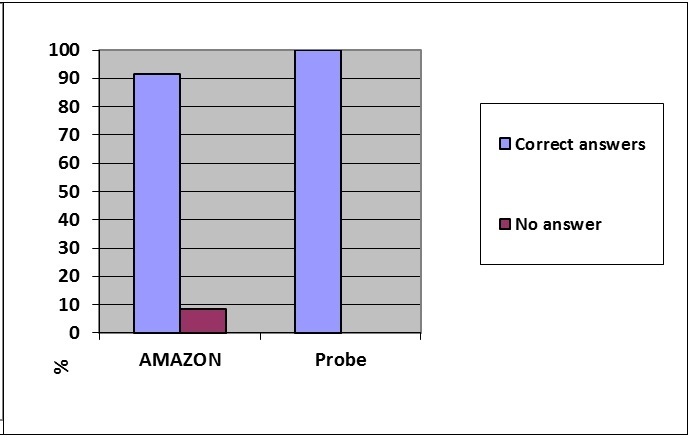}\caption{\label{fig8}Results for Question2: left) mean time for answering
(in seconds), right) number of responses within one minute}
\end{figure}

\medskip{}

\textit{Q2: \textquotedblleft{}In how many films have Martin Scorsese
and Leonardo Di Caprio acted together?\textquotedblright{} }

The mean times under the two environments for answering is presented
in Figure \ref{fig8} with only 11 measures since one of the subjects
failed to provide an answer with Amazon within the minimum delay of
one minute. It seems that the probe clearly outperforms Amazon with
nearly half mean time (confidence for H0: p =1.2E-07) 

In fact the difference of mean value is not as instructive as it may
look in this experiment; some testers took time for answering, particularly
with Amazon, because they knew they had one minute and they did not
want to give wrong answers. However this testers\textquoteright{}
strategy applied for both applications and the difference of mean
values is still interesting. The most interesting result is that one
tester failed to find the answer with Amazon when all testers succeeded
with the probe (it must be noticed that this tester\textquoteright{}s
failure is not taken into account in the mean time for the benefit
of Amazon). 

\medskip{}
\textit{Q3: \textquotedblleft{}Here are five actors: Matt Damon, Al
Pacino, Julia Roberts, Brad Pitt and Georges Clooney. In how many
films have they acted \ldots{} }

\textit{- \ldots{} together }

\textit{- \ldots{} four among the five }

\textit{- \ldots{} three among the five }

\textit{- \ldots{} two among the five?\textquotedblright{} }

\begin{figure}[h]
\centering{}\includegraphics[width=5.4cm]{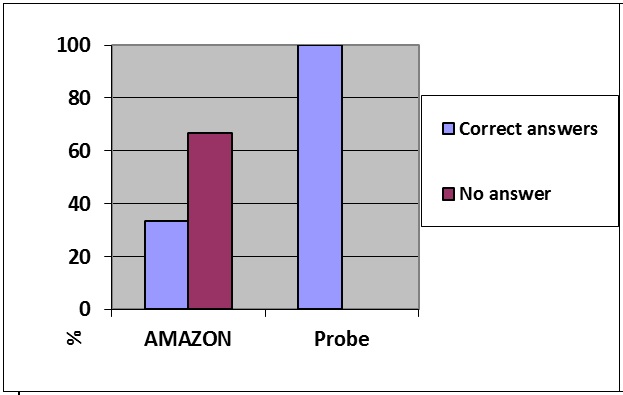}\includegraphics[width=5.5cm]{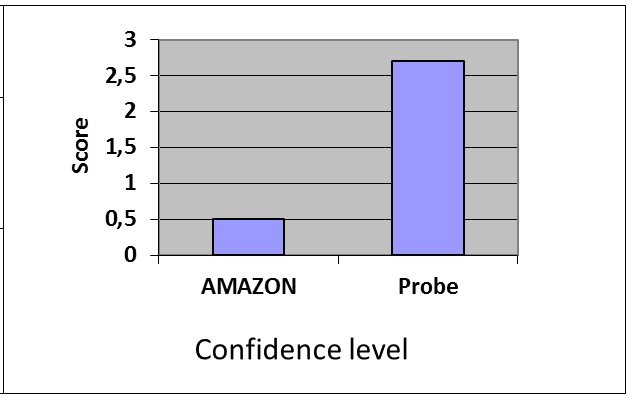}\caption{\label{fig9}Results for question 3: left) number of answers, right)
confidence degree}
\end{figure}

For this complex question involving more semantics, figure \ref{fig9}
left shows that only one third of the subjects could provide answers
using Amazon within a two minute timeout whilst all answers were provided
with the probe. Interestingly, for those who gave an answer, the degree
of confidence on a scale of 0 to 3 was low under Amazon and high with
the probe (see figure \ref{fig9} right). t- test cannot be applied
because a majority of testers failed to give an answer within the
time delay. 

The fourth question concerned the capacity of combining two semantic
view points.

\medskip{}
\textit{Q4: \textquotedblleft{}You want to go to the cinema with a
friend. You like Brad Pitt and Georges Clooney and your friend likes
Julia Roberts and Brad Pitt. What is the best choice for you, for
her and the best compromise for both?\textquotedblright{} }

\begin{figure}[h]
\centering{}\includegraphics[width=4.5cm]{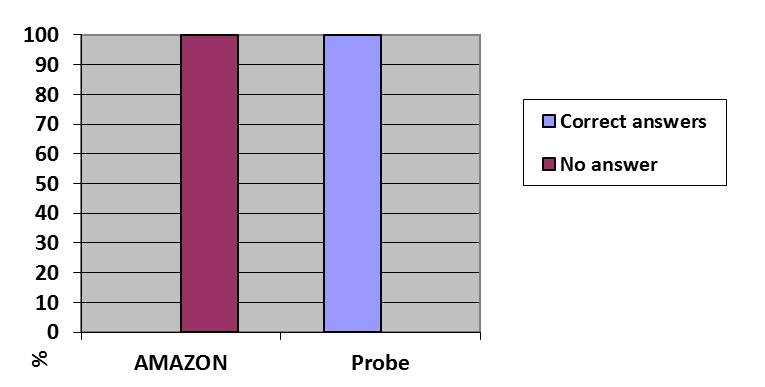}\caption{\label{figure10}results for question 4}
\end{figure}

The results are particularly interesting. No one was able to give
any answers with Amazon in less than 2 minutes (see figure \ref{figure10})
whilst all answers were given with the probe in a mean time of 64.9
seconds.

\begin{figure}[h]
\centering{}\includegraphics[width=4.5cm]{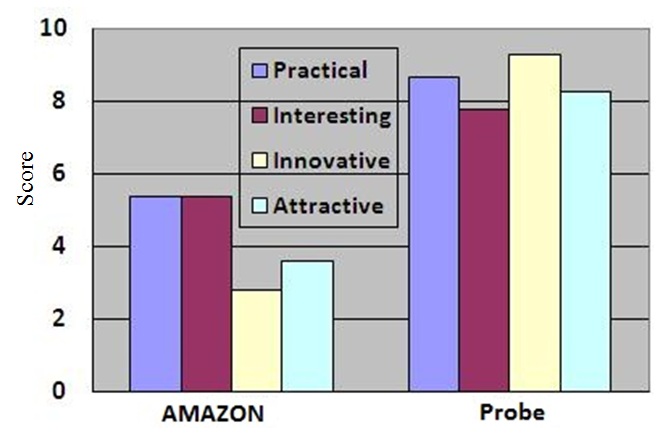}\caption{\label{figure11}Subjective results for question 5: Satistfaction
criteria}
\end{figure}

Figure \ref{figure11} summarises the answers regarding practicality,
interest, innovation and enjoyment. Each method is assessed independently
by the subjects. The semantic probe method clearly outperforms the
more traditional Boolean search method with statistically significant
results for all four answers (p \textless{} 0.001). These results
are particularly interesting when going back to those detailed in
\citet{Godin95incrementalstructuring} where a similar test was conducted
comparing a query based search with a local Hasse diagram driven navigation
search. The authors report equal performances whilst the tests we
conducted with semantic probes give much better performances. 

The last question was a key assessment.

\medskip{}
\textit{Q6: \textquotedblleft{}If you had to choose between the two
methods, which one would you prefer?\textquotedblright{} }

\begin{figure}[h]
\centering{}\includegraphics[width=4.5cm]{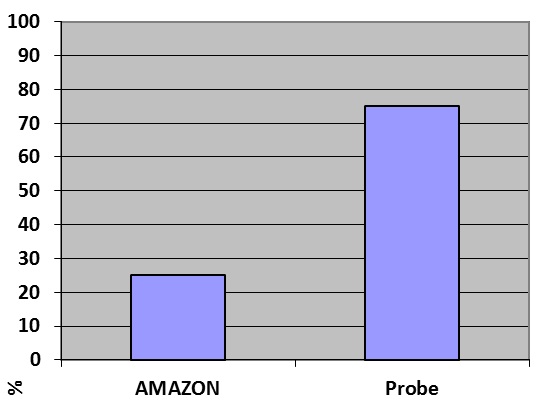}\caption{\label{figure12}Results for question 6}
\end{figure}

Figure \ref{figure12} shows the answer to the question. Nine subjects
favoured the probe. Three subjects preferred the traditional Boolean
search and its list presentation although they favoured the probe
when answering question 5. They were asked the reason for this contradiction.
They had the same answer. They were used to buying music or DVDs online
and did not expect more from the Internet. They were not concerned
with more semantically sophisticated methods as they found they had
no use for them. 

\medskip{}
\textbf{Analysis of results. }

\begin{figure}[h]
\centering{}\includegraphics[width=4.5cm]{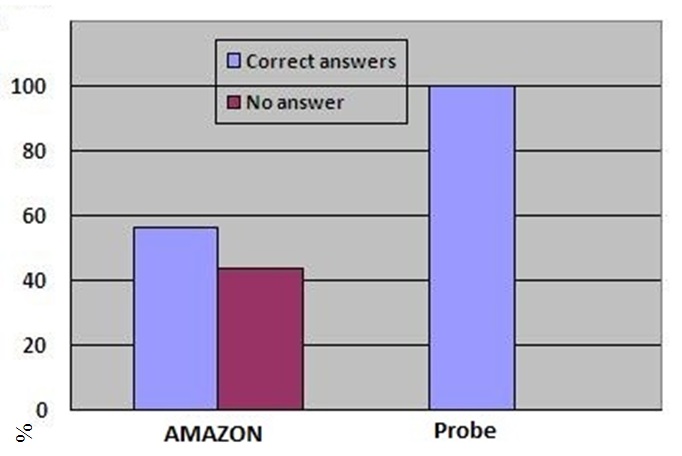}\caption{\label{figure13}Subjective results}
\end{figure}

\medskip{}
Synthesis of results for the 4 first questions and all 12 testers,
i.e. 60 answers is presented in figure \ref{figure13}. This figure
demonstrates that all questions could be answered with the semantic
probe within the time limit whilst nearly half could not be answered
using Amazon. In particular, nobody was capable to answer complex
questions with Amazon in the time period. When there was an answer
given, it was 10\% faster for simple questions and 50\% faster for
more complex questions when using the probe. Moreover testers' confidence
in answers is low for complex questions with Amazon and high using
probes. For instance the mean degree of testers' confidence in question
3 is 2.7 with probes and 0.5 with Amazon on a scale from 0 to 3. 

This set of simple tests on a medium size database shows on the one
hand that navigating on a Hasse diagram is difficult for lay users,
and on the other hand that the semantic probe outperforms traditional
Boolean querying on a standard engine. Satisfaction questions (Q5)
confirm these conclusions: the semantic probe is favoured. These results
are obtained with a limited number of questions and a limited number
of subjects. Our intention was to set up a wider test with more subjects.
However there was an important objection to this idea. Answers to
the last questions Q5 and Q6 show that there is a contradiction between
indoor tests and reality. Conducting other more significant tests
would probably confirm the first results and would not be interesting.
Conversely since \textquotedblleft{}a common evaluation measure for
any technology is adoption by others, and the move into commercial
products\textquotedblright{} (\citet{Plaisant2004}) it was decided
that the second phase of tests should favour beta testing on real
data and move to industrial judgement through presentations and marketing.
Moreover considering the hypothesis that users would not adopt a better
technology if it does not bring about new useful functional novelty,
we explored extended services with some industrials.

\subsection{Second phase}

\textbf{Controlled experiments with real data}

\medskip{}
Another set of controlled experiments was intended to test the semantic
probe approach on data with real users in a real environment. We downloaded
three users\textquoteright{} photo album from their Facebook profiles,
and asked them to navigate in their respective albums looking for
photos with some of their friends. Only tagged photos were loaded.
We took account with rather big amount of photos, between 1000 and
1200 tagged photos in each album and more than 250 friend tags, nearly
8 times the size of our DVD benchmark. We asked similar questions
as those for the DVD benchmark. We added more difficult questions
such as \textquotedblleft{}find the photos with the most people\textquotedblright{}.
Our intention was to assess scalability on real medium size databases,
the interest of users and the ease of use. Results on scalability
were excellent (speed and reactions to interactions). Ease of use
was as expected: looking for a particular friend took less than 10
seconds with the probe and at least 20 seconds on Facebook. 

Some experiments were not possible with Facebook when they were easy
with the semantic probe, like finding photos with three particular
persons. Tester\textquoteright{}s interest was high when they rediscovered
their photos and confessed they did not know of any applications that
could provide these functionalities. 

Figure \ref{figure15} shows a screen capture taken during testing.

Testers' were asked questions simular to Question 4 : \textit{\textquotedblleft{}You
want photos displaying people that are both known to you and a friend.
You like person A and Person B and your friend likes person C and
Person A. What is the best choice for you, for her and the best compromise
for both?\textquotedblright{}}. The results are particularly interesting.
No one was able to give any answers with Facebook in less than 2 minutes,
whilst all answers were given with the probe in an acceptable time.

\begin{figure}[h]
\centering{}\includegraphics[width=10cm]{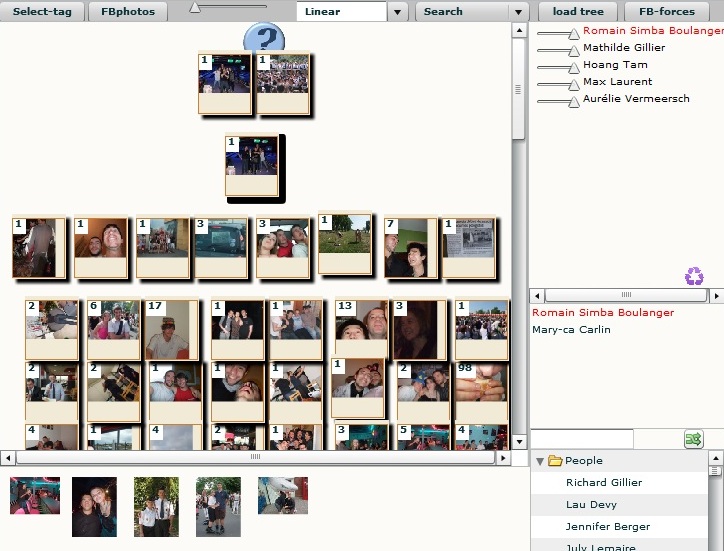}\caption{\label{figure14}Snapshot of a test with 1288 photos from Facebook}
\end{figure}

\begin{figure}[h]
\centering{}\includegraphics[width=10cm]{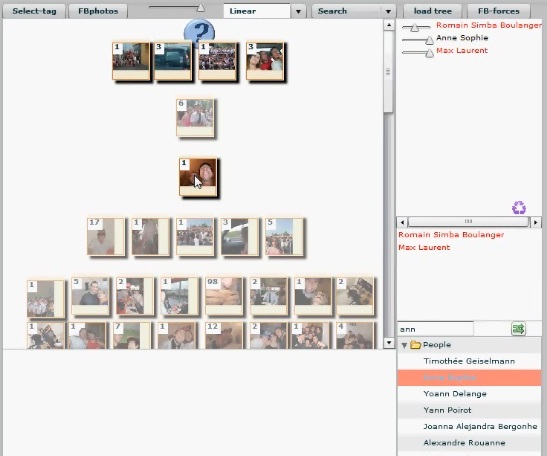}\caption{\label{figure15}Snapshot of a test with weighted concepts from Facebook
photo collection}
\end{figure}

Figure \ref{figure14} shows a screen capture taken during testing.

One of the testers spent about half an hour, using the probe to explore
the photo set, and search for photos with his friends.

Moreover, one student having been informed of the probe tool by one
of his friend asked to use the tool, to explore his photo albums and
sort his collection using the semantic probe. He found the tool very
useful to browse photos and make complex search tasks.

He was continuously tempted to add photos on the probe an see the
computed photo sub-hierarchy, helping him to browse his collection
and rediscover his group photos, and the events associated with the
photos.

He reported a situation of dropping on to the probe a person and after
viewing the photos, he tooks one of them with persons of interest
for him and drop this photo on to the probe. He then discovered very
quickly (less than a second) other photos with these persons in a
hierarchic tree containing photos with one or more of these persons.
He was then able to explore these photos by hovering the mouse on
them and seen the persons present on these photos. He then navigate
quite a long time putting photos on the probe and exploring his collection. 

In Figure \ref{figure15} this user associated weights to some persons,
seen on the upper right zone of the image, in order to favor photos
with his best friend while keeping other friends on the photos. The
subhierarchy of photos then reorganized according to the given weights. 

Several users asked if it was possible to use the probe directly integrated
in Facebook, and others asked us when this new tool will be available.

These conclusions confirmed that industrial applications should be
considered. 

\medskip{}
\textbf{Deployment: Industrial assessment and applications }

\begin{figure}[h]
\centering{}\includegraphics[width=10cm]{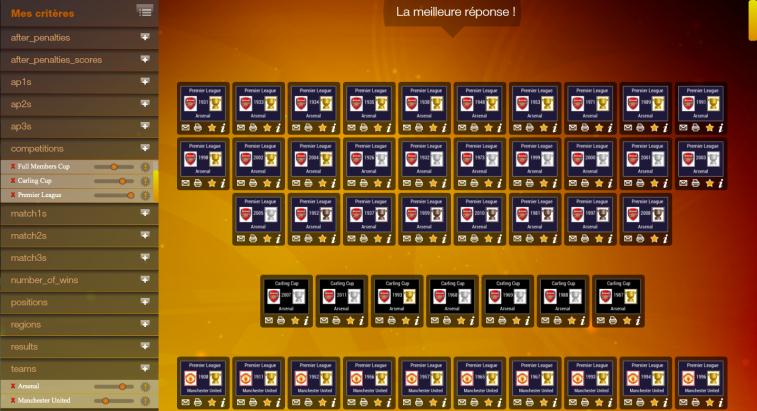}\caption{\label{figure16}Experimental interface for a sports TV channel}
\end{figure}

\medskip{}
After patenting, a license agreement was signed with a software distributor.
Many presentations have then been given by our partner to industrials
in Europe and in the U.S with excellent feedbacks. Several prototypes
are now under development with redesigned interfaces for each industrial
target such as a pharmaceutical world leader (drug interactions),
TV channels (programs selection), a music major company, a human resource
management company, etc. Figure \ref{figure16} shows an experimental
interface for a sports TV channel in the U.K. Performances between
soccer clubs (in this case Arsenal and Manchester\textquoteright{}s
victories in competitions) can be distinguished through weighted criteria
(on the left). Optimisation allows now the management of thousands
of objects, still keeping the display simple, attractive and conceptually
rich, far from simple lists or grids. A commercial application is
now installed in a show room in Casablanca. This important industrial
feed-back shows that we have partly reached our goal of bringing Galois
lattices from experts to novices. At the time of writing this paper
it is not yet possible to tell how many of these possibilities will
turn out into commercial applications due to other industrial considerations
than mere functional innovation or aesthetic interest. Industrials
are still hesitating to invest in a new technology, whatever its interest,
when it competes with well established simple technologies, unless
it brings about new interesting services. This is confirmed by the
contradiction we observed in the answers of Questions 5 and 6: although
the probe was clearly preferred, three testers would continue to use
Amazon because they were used to it. It may be a minority of people,
but this minority represents industrial reality. To overcome this
difficulty a new technology must offer more. This is what we now explore.
The probe approach should offer users a new insight in their data. 

\medskip{}
\textbf{New services: indexing and complementarity}

\medskip{}
We already know from \citet{10.1109/TVCG.2009.201} that a semantic
probe is a good means for indexing objects with objects. This is an
important new functionality which we have improved a lot but which
is not industrially explored yet. Another functionality is shown in
this paper because it is particularly interesting in many domains
and unveils new insights in databases. The probe can be used for searching
for complementary data or objects.

\begin{figure}[h]
\centering{}\includegraphics[width=10cm]{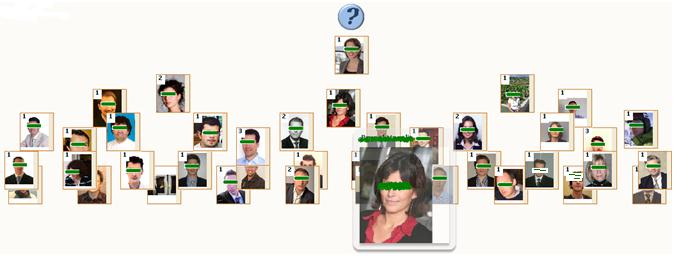}\caption{\label{figure17}Experimental interface for a company with a hundred
people tagged with their competences}
\end{figure}

It was drafted by our software partner for the human resource department
of a big international electro-mechanics leader company. People from
the company are tagged with their competences from a thesaurus; their
location and their availability are also defined as properties. Figure
\ref{figure17} shows a snapshot of the mock-up whose database contains
a hundred people (eyes and names are barred in this paper and tags
are not shown for obvious privacy and confidential reasons). Another
interesting feature of this mock-up is that the probe\textquoteright{}s
tags are weighted through the use of sliders presented in Figure \ref{figure15}
and \ref{figure16}. When looking for a particular profile for a project,
a human resource manager can load the probe with the expected competences
and data. There will hopefully be some people meeting the criteria
like the woman just under the probe in the figure. However it is more
interesting to see how a team of people can be built from different
people\textquoteright{}s competences. Groups differ according to some
competences. Those that come up next to the probe partly meet the
requirements. The union of their extents (comptences) may lead to
a super-group whose extent matches the probe\textquoteright{}s profile.
This new functionality which was suggested by our industrial visitors
leads us to consider complementary concepts, i.e. relations between
different concepts which may merge for meeting some overall requirements.
This mock-up opens up new research directions for knowledge mining,
complementary social networks and information visualization.

\subsection{Scalability issues }

One of the main arguments in favour of the semantic probe paradigm
is its visual scalability for novices compared to Hasse Diagrams.
Consequently it is important to analyse how scalability impacts onto
semantic probe displays. In a normal Galois lattice, the number of
concepts upper bound equals 2N for N attributes. In practice it is
never the case. First we must consider the number of groups because
strictly similar attributes always regroup in the same concepts. Second
it is shown in \citet{Godin:1993:ECN:172206.172208} that if K is
a fixed upper bound on the number of attributes for each group, the
number of nodes \textbar{}H\textbar{} in the Hasse diagram is bounded
by $2^{K}n$ where n is the number of attributes. Third these authors
also show that in real cases because of the attributes\textquoteright{}
repartition this upper bound is between $4\times n$ to $11\times n$.
In the semantic probe case we only display groups and concepts appear
as a result of interaction. The upper bound of visible groups is equal
to n if the probe were loaded with all objects, which is already 4
times to 10 times less than the number of nodes of a Hasse diagram.
In practice, the probe should be loaded with few objects and a little
proportion of groups should be visible. In normal usage, scalability
is not a problem for semantic probes. Experimentation confirms these
results: the Facebook photo tests involved more than 1000 items and
most of industrial prototypes involve more than 10000 items.

Beyond this quantitative consideration, the most important visual
complexity reduction factor is the absence of edges. Edge crossing
is a key problem in graph drawing particularly in the case of object-attribute
sets. It is known that a graph containing at least a 5 node clique
($K_{5}$) is not planar and necessarily opens up an essential problem
of edge crossing visual difficulty. In general Hasse diagrams are
far beyond this limit. Probe driven sub-hierarchies ignore this problem.
This visual simplification explains the good performances in the controlled
experiments and the welcoming by end users and industrials.

\section{CONCLUSION }

Irrespective of what visualization strategy is employed, it is difficult
to display object-attribute databases with their topological properties.
Galois lattices are good at solving this problem through the use of
Hasse diagrams. They provide a powerful tool for knowledge analysis
but they fall short from addressing the complexity and scalability
bottlenecks for novices. 

We proposed an interactive user-centric probe-driven strategy. Our
results confirm that this approach, although it does not replace existing
ones, improves navigation and is attractive for industrial partners
in varying fields. However the issue of providing conceptually enhanced
visualization solutions to users at the expense of user acceptance
is still on-going. Simple experiments and hesitation among interested
industrials show that a new technology must outperform in many ways
a simple established technology to become attractive. This is why,
although our probe approach shows many qualities, we consider that
new services must be provided to reach industrial applications. Some
promising experiences are being performed in this direction with assistance
to indexing and the original idea of data complementarities.

\section{References}

\bibliographystyle{plainnat}
\bibliography{19D__MP_RECHERCHE_bibtec_infovis}

\end{document}